\documentclass{article}
\usepackage{spconf,amsmath,graphicx}
\usepackage{todonotes}
\usepackage{eso-pic}
\usepackage{lipsum} 

\newcommand{\copyrighttext}{%
  \footnotesize{%
    \textcopyright{} 2023 IEEE. Personal use of this material is permitted. Permission from IEEE must be obtained for all other uses, in any current or future media, including reprinting/republishing this material for advertising or promotional purposes, creating new collective works, for resale or redistribution to servers or lists, or reuse of any copyrighted component of this work in other works.%
  }%
}

\AddToShipoutPictureBG*{%
  \AtPageLowerLeft{%
    \put(55,25){%
      \parbox{1.0\textwidth}{%
        \centering\copyrighttext%
      }%
    }%
  }%
}

\newcommand{\model}{\texttt{LGAP}}

\title{Language Guided Adversarial Purification}
\name{Himanshu Singh, A V Subramanyam}
\address{Indraprastha Institute of Information Technology, Delhi, India}
\begin{document}
%
\maketitle
\begin{abstract}
Adversarial purification using generative models demonstrates strong adversarial defense performance.
These methods are classifier and attack-agnostic, making them versatile but often computationally intensive. Recent strides in diffusion and score networks have improved image generation and, by extension, adversarial purification. Another highly efficient class of adversarial defense methods known as adversarial training requires specific knowledge of attack vectors, forcing them to be trained extensively on adversarial examples. To overcome these limitations, we introduce a new framework, namely Language Guided Adversarial Purification (\model), utilizing pre-trained diffusion models and caption generators to defend against adversarial attacks. Given an input image, our method first generates a caption, which is then used to guide the adversarial purification process through a diffusion network. Our approach has been evaluated against strong adversarial attacks, proving its effectiveness in enhancing adversarial robustness. Our results indicate that \model\ outperforms most existing adversarial defense techniques without requiring specialized network training. This underscores the generalizability of models trained on large datasets, highlighting a promising direction for further research.
\end{abstract}

\section{Introduction}
\label{sec:intro}

The use of deep neural networks, especially within the realm of computer vision, has ushered in transformative advancements in various applications. Despite these strides, a consistent vulnerability is the susceptibility of such models to adversarial perturbations \cite{goodfellow_explaining_2015}. These perturbations, often imperceptible, can fool even the most sophisticated neural networks, causing them to misclassify inputs. Addressing this alarming vulnerability has become a research imperative, leading to a rapidly growing body of literature dedicated to understanding and defending against these adversarial threats \cite{madry2018towards, song2018pixeldefend}.

\begin{figure}[t]
    \centering
    \includegraphics[scale=0.90]{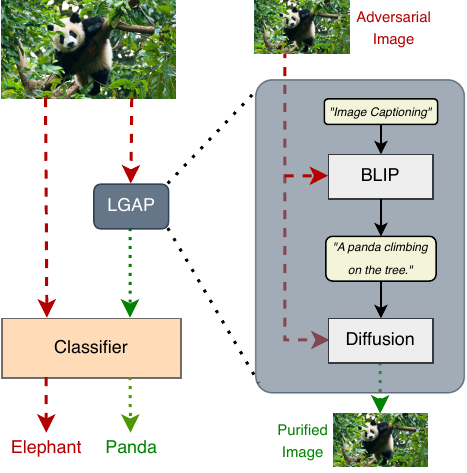}
    \caption{Illustration of \model. A pre-trained image-captioning model (BLIP) generates captions for input images, providing a textual representation of the visual content. Leveraging the generated captions,  purified images are created via the diffusion model. The red dashed lines represent the adversarial image input, while the green dotted lines indicate the resulting purified image.}
    \label{fig:model}
\end{figure}

Historically, adversarial training, introduced by Goodfellow \textit{et al.} \cite{goodfellow_explaining_2015}, has been posited as an effective defense strategy. This approach, which integrates adversarial examples into the training phase, aims to strengthen models against specific adversarial attacks. However, its efficacy is often limited to the spectrum of attacks encountered during training, thereby leaving models vulnerable to novel adversarial strategies. This constraint underscores the necessity for alternative defensive paradigms.

Given their inherent capability to generate or transform data, generative models have recently been explored as potential tools for adversarial purification \cite{samangouei2018defense, shi2020online, yoon2021adversarial}. Within this domain, diffusion models have emerged as particularly promising candidates. 
Recent studies, as exemplified by Nie \textit{et al.} \cite{nie_diffusion_2022} and Carlini \textit{et al.} \cite{carlini2022certified}, have harnessed the potential of score-based and diffusion models towards purification of adversarial samples.  

Primarily, the adversarial purification techniques have focussed only on the image modality, despite promising performance of diffusion models in multi-modal tasks such as text-to-image generation \cite{rombach_high-resolution_2022}. Thus, in our work, we investigate the impact of language towards the robustness of vision models.
Our research focuses on defensive strategy based on vision and language models trained on large datasets. By leveraging the capabilities of such models trained jointly on language and vision tasks, we propose a novel framework of \textbf{L}anguage \textbf{G}uided \textbf{A}dversarial \textbf{P}urification (\model), as illustrated in Figure \ref{fig:model}. This novel framework, which seamlessly integrates a caption generator and a pre-trained diffusion model with a classifier, leverages the inherent generalisability of these models to purify an adversarial input. To the best of our knowledge, language based adversarial purification has not been addressed in the literature.

We conduct elaborate empirical evaluations across benchmark datasets, including ImageNet \cite{imagenet_cvpr09}, CIFAR-10 \cite{krizhevsky2009learning} and CIFAR-100 \cite{krizhevsky2009learning}. The results of evaluation against $L_\infty$ norm attacks  corroborate the robustness of our framework. Notably, for the ImageNet, our method reveals  better performance compared to previous techniques.

\section{Related Works}
\label{sec:relatedworks}

\textbf{Diffusion models in image generation:} The landscape of image generation has been revolutionized by diffusion models. Rooted in the foundational works of Sohl-Dickstein \textit{et al. }\cite{sohl2015deep} and later extended by Song \textit{et al.} \cite{song2019generative} and Ho \textit{et al.} \cite{ho2020denoising}, these models have exhibited unparalleled prowess in generating high-quality image samples. Song \textit{et al.} \cite{song2020score} further advanced this domain by combining generative learning mechanisms with stochastic differential equations, thereby broadening the horizon of diffusion models. 

\noindent \textbf{Language-image pretraining:} A significant milestone in deep learning, language-image pretraining bridges the gap between textual and visual data. Pioneering models such as CLIP \cite{radford2021learning} and BLIP \cite{li2022blip} have leveraged vast amounts of text and image data to jointly train vision and language models, demonstrating tremendous progress in multi-modal tasks. 

\noindent \textbf{Adversarial training:}
The foundational work of Madry \textit{et al.} \cite{madry2018towards} established adversarial training as a robust method for safeguarding neural networks from known adversarial attacks. While the effectiveness of the method is well-recognized, its scalability and adaptability have been enhanced through inspirations from metric learning \cite{mao2019metric} and self-supervised paradigms \cite{chen2020self}. However, the computational demands of adversarial training has spurred research into more efficient training methods \cite{shafahi2019adversarial,wong2019fast}.

\noindent\textbf{Adversarial purification}: Generative models have emerged as a pioneer in the adversarial purification realm. Initial endeavors, such as those by Samangouei \textit{et al.} \cite{samangouei2018defense}, harnessed GANs for purification. Subsequent innovations leaned on energy-based models (EBMs) to refine the purification process using Langevin dynamics \cite{grathwohl2019your}. Notably, the intersection of score networks and diffusion models with adversarial purification has been explored recently, with promising results against benchmark adversarial attacks \cite{yoon2021adversarial, nie_diffusion_2022}.

\section{Proposed Method}
\label{sec:proposed}

We propose a novel defense strategy against adversarial attacks on classification models by leveraging language guidance in diffusion models for adversarial purification. For a clean sample \( \mathbf{x} \) with label \( y \), and a target neural network \( f_{\boldsymbol{\theta}} \), the adversary aims to produce \( \mathbf{x}_{\text{adv}} \) by introducing adversarial perturbations. This results in a prediction \( f_{\boldsymbol{\theta}}(\mathbf{x}_{adv}) \) that differs from the original prediction \( f_{\boldsymbol{\theta}}(\mathbf{x}) = y \). The underlying premise of the proposed method is to preprocess the input \( \mathbf{x} \) through a diffusion model conditioned on a caption to remove any adversarial perturbations before feeding it to \( f_{\boldsymbol{\theta}} \). We first discuss the caption generation followed by purification using diffusion model.

\subsection{Image captioning}
For image captioning, we use a caption generator from BLIP \cite{li2022blip}. 
BLIP has a multi-modal encoder-decoder architecture which consists of three major components a unimodal encoder for generating image and text embeddings, an image-grounded text encoder that computes cross attention and self-attention between the two encodings to give a multimodal representation of image text pair, and an image-grounded text decoder that uses casual self-attention to give the text caption.
We use the unimodal encoder and image-grounded text decoder to generate the captions. Given an input $\mathbf{x}$, the captions are generated as, 
\begin{equation*}
\text{\textit{Caption}}_{\text{BLIP}} = \text{\textit{BLIP}}(\mathbf{x}).
\end{equation*}

We show some sample captions in Figure \ref{fig:samples}. We can see that the captions for the clean samples (top row) contains the true label. In the second row, adversarial samples are given and the classifier' prediction is incorrect. Here, \textit{truck} is classified as \textit{ship}. However, the BLIP caption still contains the true label \textit{truck}, though the caption is not the same as that of clean sample. Thus, using these captions can condition the diffusion models with true semantics which can enhance purification of the adversarial images. Next, we discuss the diffusion based purification.

\subsection{Diffusion purification process}

    

\noindent \textbf{Latent diffusion process}
In a standard diffusion model \cite{ho2020denoising}, the diffusion process can be defined as:
\begin{equation*}
{\bf{x}_t} = \sqrt{1 - \beta_t} \cdot {\bf{x}_{t-1}} + \sqrt{\beta_t} \cdot \boldsymbol{\epsilon}_t
\end{equation*}
where \( \beta_t \in (0,1) \) is the variance schedule, ${\bf{x}_t}$ is the noisy sample, and $\boldsymbol{\epsilon}_t$ is the noise at time step \( t \).
In Latent Diffusion Models \cite{rombach_high-resolution_2022}, this process is applied in latent space:
\begin{gather*}
\mathbf{z}_0 = \mathcal{E}(\mathbf{x})\\
{\bf{z}_t} = \sqrt{1 - \beta_t} \cdot {\bf{z}_{t-1}} + \sqrt{\beta_t} \cdot {\boldsymbol{\epsilon}_t}
\end{gather*}
where $\mathbf{z}_0$ is the latent vector obtained from the encoder $\mathcal{E}$ and $\mathbf{z}_t$ is the noisy latent vector at time step \(t\). 

\noindent \textbf{Reverse process in latent space}
In the reverse process, the aim is to recover $\mathbf{z}_0$ from $\mathbf{z}_T$ given a sequence of noise terms $\boldsymbol{\epsilon}_t$. Mathematically, this is defined as:
\begin{equation*}
{\mathbf{z}_t} = g_{\theta}({\mathbf{z}_{t+1}}, t, {\boldsymbol{\epsilon}_t})
\end{equation*}
where ${g_{\theta}}$ is a parameterized model. Additionally, ${g_{\theta}}$ is conditioned on text by augmenting ${g_{\theta}}$ architecture with cross attention layers. Since our goal is to leverage the BLIP generated captions, we condition the diffusion model as:
\begin{equation*}
{\mathbf{z}_t} = g_{\theta}({\mathbf{z}_{t+1}}, t, {\boldsymbol{\epsilon}_t}, \mathbf{C})
\end{equation*}
where $\mathbf{C} = {\tau_{\theta}}(\text{\textit{Caption}}_{\text{BLIP}})$, and \( \tau_{\theta} \) is text encoder. Since BLIP is a powerful model, the likelihood that it correctly identifies the image is high. This gives a better guidance to diffusion model compared to image-only case.


\noindent \textbf{Final image reconstruction and training}
Finally, the reconstructed image $\hat{\mathbf{x}}$ can be obtained from the reconstructed latent representation ${\mathbf{z}}_0$ as, $\hat{\mathbf{x}} = \mathcal{D}({\mathbf{z}}_0)$, where $\mathcal{D}$ is the decoder.
Given model  $f_{\boldsymbol{\theta}}$, clean image $\mathbf{x}$, its corresponding pre-processed sample $\hat{\mathbf{x}}$ and labels $y$, we optimize for,
$\arg \min_{\boldsymbol{\theta}}\frac{1}{n}\sum_{i=1}^n \mathcal{L}_{CE}(f_{\boldsymbol{\theta}}(\hat{\mathbf{x}}_i), {y}_i)$, 
where $\mathcal{L}_{CE}$ is the cross-entropy loss and $n$ is the number of samples. 

In contrast to adversarial training of several epochs with adversarial samples, we only need a few epochs of finetuning with pre-processed clean samples. Further, compared to score or diffusion-based purification, which extensively trains these models, we only need minimal training of the classifier. 


\section{Experiments and Results}
\label{sec:results}
\subsection{Experimental settings}

\textbf{Datasets and network architectures:} Our experimental evaluation involves three datasets, namely CIFAR-10 \cite{krizhevsky2009learning}, CIFA-100 \cite{krizhevsky2009learning} and ImageNet \cite{imagenet_cvpr09}. We utilize the base models from RobustBench \cite{croce2020robustbench} model zoo for CIFAR-10 and ImageNet. For CIFAR-100 we train the model following Yoon et al. \cite{yoon2021adversarial}. We compare our approach against other adversarial purification strategies on CIFAR-10, adhering to their experimental configurations. We also evaluate our method against preprocessor blind attacks on ImageNet. Regarding classifier architectures, we opt for two prevalent models: ResNet-50  \cite{he2016deep} for ImageNet and WideResNet-28-10 \cite{zagoruyko2016wide} for CIFAR-10 and CIFAR-100. We fine-tune the WideResNet on images generated from the diffusion network for 15 epochs. We utilize Adam optimizer with a $10^{-3}$ learning rate. For generating captions, we use pre-trained BLIP \cite{li2022blip} with default hyperparameters, and for the diffusion process, we use a pre-trained latent diffusion model from \cite{rombach_high-resolution_2022} with default parameters except for the noise parameter \textit{t}. We set \textit{t} to 0.5 for CIFAR-10, and CIFAR-100 and 0.1 for ImageNet. We will be releasing the code soon.

\begin{figure}[t]
    \centering
    \includegraphics[scale=0.75]{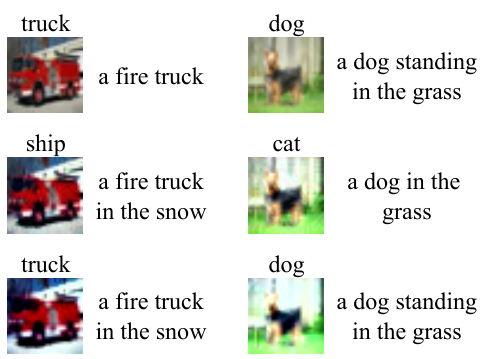}
    \caption{Purified samples given by \model. The first, second, and third rows contain clean, adversarial, and purified samples. The BLIP generated captions are given on the right, and the predicted label is on top of the image.}
    \label{fig:samples}
\end{figure}

\begin{table}[]
\small
\centering
\begin{tabular}{lccc}
\hline
\multicolumn{1}{c}{\textbf{Methods}}    & \multicolumn{2}{c}{\textbf{Accuracy(\%)}} & \textbf{Architecture} \\
                              & Natural          & Robust   &              \\ \hline
Raw WideResNet                & 95.80             & 0.00     & WRN-28-10    \\
Adv. purification methods      &            &          &              \\
\textbf{LGAP}                          & 90.03             & 71.68    & WRN-28-10    \\
Yoon et al. \cite{yoon2021adversarial}*     &   &   &\\
\qquad (\( \sigma \) = 0.1)*                 & 93.09             & \textbf{85.45}    & WRN-28-10    \\
\qquad (\( \sigma \) = 0.25)*                & 86.14             & 80.24    & WRN-28-10    \\

Hill et al. \cite{hill2020stochastic}*               & 84.12             & 78.91    & WRN-28-10    \\
Shi et al. \cite{shi2020online}*           & \textbf{96.93}             & 63.10    & WRN-28-10    \\
Du et al. \cite{du2019implicit}*        & 48.7              & 37.5     & WRN-28-10    \\
Grathwohl et al. \cite{grathwohl2019your}*      & 75.5              & 23.8     & WRN-28-10    \\
Song et al. \cite{song2018pixeldefend}*          &                   &          &              \\
\qquad Natural + PixelCNN            & 82                & 61       & ResNet-62    \\
\qquad AT + PixelCNN                 & 90                & 70       & ResNet-62    \\ \hline
Adv training methods    &          &              \\
Madry et al. \cite{madry2018towards}*         & 87.3              & 70.2     & ResNet-56    \\
Dong et al. \cite{dong2023enemy}*            & 84.98                       & 51.29                      & ResNet-18             \\ \hline
\end{tabular}
\caption{Results for preprocessor blind PGD attack for CIFAR-10, within an \( L_{\infty} \) \( \epsilon \)-ball, where \( \epsilon \) = {8}/{255}. Data sourced from existing literature is indicated by an asterisk *.}
\label{tab:pp-blind}
\end{table}

\noindent\textbf{Adversarial attacks:} We test our algorithm against preprocessor blind PGD attacks, in which the adversary has complete visibility into the classifier but is uninformed about the purification model. 
We also evaluate our algorithm against strong adaptive attack, which involves more complex scenarios due to our purification algorithm's iterative nature through neural networks, potentially leading to obfuscated gradients. To rigorously test our defense mechanism, we use potent adaptive attacks, such as Backward Pass Differentiable Approximation (BPDA) \cite{athalye2018obfuscated} and its variations. We experiment with the basic form of BPDA, where the purification function is approximated as the identity function. We further validate its robustness using Expectation Over Time (EOT) attacks \cite{athalye2018obfuscated}.

\begin{table}[]
\small
\centering
\begin{tabular}{lcc}
\hline
\multicolumn{1}{c}{\textbf{Method}}   & \multicolumn{2}{c}{\textbf{Accuracy}} \\
                             & Natural       & Robust      \\ \hline
\textbf{LGAP}                         & 58.71          & 39.82       \\
Yoon et al.\cite{yoon2021adversarial}*                  & \textbf{77.83}          & \textbf{43.21}       \\ \hline
Adversarial training methods &                &             \\
Madry et al.\cite{madry2018towards}*        & 59.58          & 25.47       \\
Li et al.\cite{li_defense-vae_2020}*        & 61.01          & 28.88      
\end{tabular}
\caption{Preprocessor blind PGD attack for CIFAR-100, \( \epsilon \) = \(8\)/\(255\). Data sourced from existing literature is indicated by an asterisk *.}
\label{tab:pp-blind-cifar100}
\end{table}
\begin{table}[th]
\small
\centering
\begin{tabular}{lccc}
\hline
\textbf{Methods}                           & \multicolumn{2}{c}{\textbf{Accuracy(\%)}}                         & \textbf{Architecture} \\ 
\qquad Attacks                          & \multicolumn{1}{l}{Natural} & \multicolumn{1}{l}{Robust} & \multicolumn{1}{l}{}  \\ \hline
Adv. Purification Methods & \multicolumn{1}{l}{}        & \multicolumn{1}{l}{}       & \multicolumn{1}{l}{}  \\
\textbf{LGAP}                             & \multicolumn{1}{l}{90.30}   & \multicolumn{1}{l}{}       & \multicolumn{1}{l}{}  \\
\qquad BPDA 40+EOT                      & \multicolumn{1}{l}{}        & \multicolumn{1}{l}{44.96}  & WRN-28-10           \\
\qquad BPDA                             & \multicolumn{1}{l}{}        & \multicolumn{1}{l}{62.63}  & WRN-28-10             \\
Yoon et al. \cite{yoon2021adversarial} (\( \sigma \) = 0.25)*                   &                             &                            &                       \\
\qquad BPDA 40+EOT                      & 86.14                       & 70.01                      & WRN-28-10             \\
Yoon et al. \cite{yoon2021adversarial} (\( \sigma \) = 0.0)*                    &                             &                            &                       \\
\qquad BPDA                             & 90.60                        & \textbf{76.87}                      & WRN-28-10             \\
Hill et al. \cite{hill2020stochastic}*              &                             &                            &                       \\
\qquad BPDA 50+EOT                      & 84.12                       & 54.9                       & WRN-28-10             \\
Song et al. \cite{song2018pixeldefend}*             &                             &                            &                       \\
\qquad BPDA                             & \textbf{95.00}                          & 9                          & ResNet-62             \\
Yang et al. \cite{yang2019me}*             &                             &                            &                       \\
\qquad BPDA 1000                        & 94.8                        & 40.8                       & ResNet-18             \\
\qquad (+AT. p = 0.4 -\textgreater 0.6)             & 88.7                        & 55.1                       & WRN-28-10             \\
\qquad (+AT. p = 0.6 -\textgreater 0.8) & 91                          & 52.9                       & WRN-28-10             \\
\qquad Approx. Input                    & 89.4                        & 41.5                       & ResNet-18             \\
Shi et al. \cite{shi2020online}*              &                             &                            &                       \\
\qquad Classifier PGD 20                & 91.89                       & 53.58                      & WRN-28-10             \\ \hline
Adv. training methods     &                             &                            &                       \\
Madry et al. \cite{madry2018towards}*            & 87.3                        & 45.8                       & ResNet-18             \\
Carmon et al. \cite{carmon2019unlabeled}*            & 89.67                       & 63.1                       & WRN-28-10             \\ \hline
\end{tabular}
\caption{Adaptive attacks for CIFAR-10, \( \epsilon \) = {8}/{255}.}
\label{adaptive-attacks}
\end{table}

\subsection{Comparison with state of the art}
The results for preprocessor blind setup shown in Table \ref{tab:pp-blind} on CIFAR10 show that our method gives better robust performance than most previous methods, specifically adversarial training methods, while maintaining comparable performance on natural images. Our method achieves a robust accuracy of 71.68\%, which clearly outperforms seven out of ten methods including two adversarial defense methods and five adversarial purification methods. A snapshot of adversarial samples and their corresponding purified images is given in Figure \ref{fig:samples}.

We further extend our evaluation to the CIFAR-100 dataset, with the robust performance comparisons listed in Table \ref{tab:pp-blind-cifar100}. Unlike other methods, such as the one by Yoon \textit{et al.}, which demands training a score network and noise parameter tuning, our method, LGAP delivers competitive results with substantially lower computational overhead.

Table \ref{adaptive-attacks} shows robust accuracy of our method against BPDA attack for CIFAR-10. Our method outperforms most previous techniques of adversarial purification and adversarial training. The gap in accuracy between our method and some recent techniques remains owing to other methods training the purification model on CIFAR10. Yoon \textit{et al.} and Hill \textit{et al.} which show better robust performance, train diffusion and EBM networks on CIFAR-10 for 200,000 iterations \cite{yoon2021adversarial, hill2020stochastic}. Whereas our method requires no such training.
\begin{table}[t]
\centering
\small
\begin{tabular}{lccc}
\hline
\textbf{Method}                 & \multicolumn{2}{c}{\textbf{Accuracy}} & \textbf{Architecture} \\
\qquad Attacks                 & Natural       & Robust       &              \\ \hline
Undefended             & 76.76         & 0            & ResNet-50    \\
\textbf{LGAP}                   & \textbf{69.09}         &              & ResNet-50    \\
\qquad AA                     &               & \textbf{57.12}        &              \\
\qquad BPDA-40                &               & 45.31        &              \\
\qquad PGD-10                 &               & 52.73        &              \\ \hline
Adv training methods   &               &              &              \\
Salman et al. \cite{salman2020adversarially}* & 64.02         &              & ResNet-50    \\
\qquad AA                     &               & 34.96        &              \\
Wong et al. \cite{wong2019fast}*   & 55.62         &              & ResNet-50    \\
\qquad AA                     &               & 26.24        &   
\\ \hline
\end{tabular}
\caption{Preprocessor blind attacks for ImageNet, \( \epsilon \) = \(4\)/\(255\).} 
\label{tab:imagenet-results}
\end{table}

Table \ref{tab:imagenet-results} shows the robust performance of our method for ImageNet. Due to the high computational cost of some attacks, we evaluate on a fixed set of 2048 as robust accuracy does not change much on the sampled subset compared to the whole subset \cite{nie_diffusion_2022}. We can see that even against strong adaptive attack such as BPDA-40, LGAP attains an accuracy of 45.31\% demonstrating the efficacy of the proposed method. The enhanced performance of the method can be attributed to the diffusion model trained on ImageNet. Similarly, a diffusion model trained on CIFAR-10 is expected to yield improved results when applied to CIFAR-10 classification.

\section{Conclusion}
\label{sec:conclusion}


Our method addressed key limitations in adversarial defense by introducing a language-guided purification approach. Unlike traditional methods, which require extensive computational resources and specific attack knowledge, our method leverages pre-trained diffusion models and caption generators. This reduces computational overhead and enhances scalability. Empirical tests show our approach is robust, outperforming conventional methods in several metrics, despite trailing some diffusion-based methods. Notably, this performance is achieved with minimal training and do not require adversarial samples or training the score or diffusion networks, thus broadening the method's applicability and setting a new efficiency standard. Our method underscores the generalizability of deep learning models trained on large datasets, pointing to avenues for future research, especially in model generalizability. 


\newpage
\small
\bibliographystyle{IEEEbib}
\bibliography{main}

\begin{thebibliography}{10}

\bibitem{goodfellow_explaining_2015}
Ian~J. Goodfellow, Jonathon Shlens, and Christian Szegedy,
\newblock ``Explaining and harnessing adversarial examples,''
\newblock in {\em ICLR}, 2015.

\bibitem{madry2018towards}
Aleksander Madry, Aleksandar Makelov, Ludwig Schmidt, Dimitris Tsipras, and Adrian Vladu,
\newblock ``Towards deep learning models resistant to adversarial attacks,''
\newblock in {\em ICLR}, 2018.

\bibitem{song2018pixeldefend}
Yang Song, Taesup Kim, Sebastian Nowozin, Stefano Ermon, and Nate Kushman,
\newblock ``Pixeldefend: Leveraging generative models to understand and defend against adversarial examples,''
\newblock in {\em ICLR}, 2018.

\bibitem{samangouei2018defense}
Pouya Samangouei, Maya Kabkab, and Rama Chellappa,
\newblock ``Defense-gan: Protecting classifiers against adversarial attacks using generative models,''
\newblock in {\em ICLR}, 2018.

\bibitem{shi2020online}
Changhao Shi, Chester Holtz, and Gal Mishne,
\newblock ``Online adversarial purification based on self-supervised learning,''
\newblock in {\em ICLR}, 2020.

\bibitem{yoon2021adversarial}
Jongmin Yoon, Sung~Ju Hwang, and Juho Lee,
\newblock ``Adversarial purification with score-based generative models,''
\newblock in {\em ICML}, 2021.

\bibitem{nie_diffusion_2022}
Weili Nie, Brandon Guo, Yujia Huang, Chaowei Xiao, Arash Vahdat, and Animashree Anandkumar,
\newblock ``Diffusion models for adversarial purification,''
\newblock in {\em ICML}, 2022.

\bibitem{carlini2022certified}
Nicholas Carlini, Florian Tramer, Krishnamurthy~Dj Dvijotham, Leslie Rice, Mingjie Sun, and J~Zico Kolter,
\newblock ``(certified!!) adversarial robustness for free!,''
\newblock in {\em ICLR}, 2022.

\bibitem{rombach_high-resolution_2022}
Robin Rombach, Andreas Blattmann, Dominik Lorenz, Patrick Esser, and Bj{\"o}rn Ommer,
\newblock ``High-resolution image synthesis with latent diffusion models,''
\newblock in {\em CVPR}, 2022.

\bibitem{imagenet_cvpr09}
Jia Deng, Wei Dong, Richard Socher, Li-Jia Li, Kai Li, and Li~Fei-Fei,
\newblock ``Imagenet: A large-scale hierarchical image database,''
\newblock in {\em CVPR}, 2009.

\bibitem{krizhevsky2009learning}
Alex Krizhevsky, Geoffrey Hinton, et~al.,
\newblock ``Learning multiple layers of features from tiny images,''
\newblock 2009.

\bibitem{sohl2015deep}
Jascha Sohl-Dickstein, Eric Weiss, Niru Maheswaranathan, and Surya Ganguli,
\newblock ``Deep unsupervised learning using nonequilibrium thermodynamics,''
\newblock in {\em ICML}, 2015, pp. 2256--2265.

\bibitem{song2019generative}
Yang Song and Stefano Ermon,
\newblock ``Generative modeling by estimating gradients of the data distribution,''
\newblock {\em NeurIPS}, vol. 32, 2019.

\bibitem{ho2020denoising}
Jonathan Ho, Ajay Jain, and Pieter Abbeel,
\newblock ``Denoising diffusion probabilistic models,''
\newblock {\em NeurIPS}, vol. 33, pp. 6840--6851, 2020.

\bibitem{song2020score}
Yang Song, Jascha Sohl-Dickstein, Diederik~P Kingma, Abhishek Kumar, Stefano Ermon, and Ben Poole,
\newblock ``Score-based generative modeling through stochastic differential equations,''
\newblock in {\em ICLR}, 2020.

\bibitem{radford2021learning}
Alec Radford, Jong~Wook Kim, Chris Hallacy, Aditya Ramesh, Gabriel Goh, Sandhini Agarwal, Girish Sastry, Amanda Askell, Pamela Mishkin, Jack Clark, et~al.,
\newblock ``Learning transferable visual models from natural language supervision,''
\newblock in {\em ICML}, 2021.

\bibitem{li2022blip}
Junnan Li, Dongxu Li, Caiming Xiong, and Steven Hoi,
\newblock ``Blip: Bootstrapping language-image pre-training for unified vision-language understanding and generation,''
\newblock in {\em ICML}, 2022.

\bibitem{mao2019metric}
Chengzhi Mao, Ziyuan Zhong, Junfeng Yang, Carl Vondrick, and Baishakhi Ray,
\newblock ``Metric learning for adversarial robustness,''
\newblock {\em NeurIPS}, vol. 32, 2019.

\bibitem{chen2020self}
Kejiang Chen, Yuefeng Chen, Hang Zhou, Xiaofeng Mao, Yuhong Li, Yuan He, Hui Xue, Weiming Zhang, and Nenghai Yu,
\newblock ``Self-supervised adversarial training,''
\newblock in {\em ICASSP}, 2020.

\bibitem{shafahi2019adversarial}
Ali Shafahi, Mahyar Najibi, Mohammad~Amin Ghiasi, Zheng Xu, John Dickerson, Christoph Studer, Larry~S Davis, Gavin Taylor, and Tom Goldstein,
\newblock ``Adversarial training for free!,''
\newblock {\em NeurIPS}, vol. 32, 2019.

\bibitem{wong2019fast}
Eric Wong, Leslie Rice, and J~Zico Kolter,
\newblock ``Fast is better than free: Revisiting adversarial training,''
\newblock in {\em ICLR}, 2019.

\bibitem{grathwohl2019your}
Will Grathwohl, Kuan-Chieh Wang, Joern-Henrik Jacobsen, David Duvenaud, Mohammad Norouzi, and Kevin Swersky,
\newblock ``Your classifier is secretly an energy based model and you should treat it like one,''
\newblock in {\em ICLR}, 2019.

\bibitem{croce2020robustbench}
Francesco Croce, Maksym Andriushchenko, Vikash Sehwag, Edoardo Debenedetti, Nicolas Flammarion, Mung Chiang, Prateek Mittal, and Matthias Hein,
\newblock ``Robustbench: a standardized adversarial robustness benchmark,''
\newblock {\em arXiv preprint arXiv:2010.09670}, 2020.

\bibitem{he2016deep}
Kaiming He, Xiangyu Zhang, Shaoqing Ren, and Jian Sun,
\newblock ``Deep residual learning for image recognition,''
\newblock in {\em CVPR}, 2016.

\bibitem{zagoruyko2016wide}
Sergey Zagoruyko and Nikos Komodakis,
\newblock ``Wide residual networks,''
\newblock {\em arXiv preprint arXiv:1605.07146}, 2016.

\bibitem{hill2020stochastic}
Mitch Hill, Jonathan Mitchell, and Song-Chun Zhu,
\newblock ``Stochastic security: Adversarial defense using long-run dynamics of energy-based models,''
\newblock in {\em ICLR}, 2021.

\bibitem{du2019implicit}
Yilun Du and Igor Mordatch,
\newblock ``Implicit generation and modeling with energy based models,''
\newblock in {\em NeurIPS}, 2019.

\bibitem{dong2023enemy}
Junhao Dong, Seyed-Mohsen Moosavi-Dezfooli, Jianhuang Lai, and Xiaohua Xie,
\newblock ``The enemy of my enemy is my friend: Exploring inverse adversaries for improving adversarial training,''
\newblock in {\em CVPR}, 2023.

\bibitem{athalye2018obfuscated}
Anish Athalye, Nicholas Carlini, and David Wagner,
\newblock ``Obfuscated gradients give a false sense of security: Circumventing defenses to adversarial examples,''
\newblock in {\em ICML}, 2018.

\bibitem{li_defense-vae_2020}
Xiang Li and Shihao Ji,
\newblock ``Defense-{VAE}: A fast and accurate defense against adversarial attacks,''
\newblock in {\em Machine Learning and Knowledge Discovery in Databases}, Peggy Cellier and Kurt Driessens, Eds. pp. 191--207, Springer International Publishing.

\bibitem{yang2019me}
Yuzhe Yang, Guo Zhang, Dina Katabi, and Zhi Xu,
\newblock ``Me-net: Towards effective adversarial robustness with matrix estimation,''
\newblock in {\em ICML}, 2019.

\bibitem{carmon2019unlabeled}
Yair Carmon, Aditi Raghunathan, Ludwig Schmidt, John~C Duchi, and Percy~S Liang,
\newblock ``Unlabeled data improves adversarial robustness,''
\newblock {\em NeurIPS}, 2019.

\bibitem{salman2020adversarially}
Hadi Salman, Andrew Ilyas, Logan Engstrom, Ashish Kapoor, and Aleksander Madry,
\newblock ``Do adversarially robust imagenet models transfer better?,''
\newblock {\em NeurIPS}, 2020.

\end{thebibliography}
\end{document}